\documentclass[conference]{IEEEtran}
\IEEEoverridecommandlockouts
\usepackage{cite}
\usepackage{comment}
\usepackage{amsmath,amsfonts}
\usepackage{booktabs}
\usepackage{multicol}
\usepackage{multirow}
\DeclareMathOperator*{\argmax}{arg\,max}
\usepackage{graphicx}
\usepackage{textcomp}
\usepackage{xcolor}
\usepackage{url}
\usepackage[a4paper, total={184mm,239mm}]{geometry}
\def\BibTeX{{\rm B\kern-.05em{\sc i\kern-.025em b}\kern-.08em
    T\kern-.1667em\lower.7ex\hbox{E}\kern-.125emX}}

\usepackage{xspace}
\usepackage{paralist}
\usepackage{subcaption}
\usepackage{bm}
\usepackage{stmaryrd}
\usepackage[noend]{algpseudocode} 
\usepackage{algorithm}
\usepackage{amssymb}
\usepackage{wrapfig}
\usepackage{colortbl}
\usepackage[switch]{lineno}

\definecolor{mygreen}{RGB}{200, 255, 200}
\newcommand{\tbgreen}{\cellcolor{mygreen}}

\newcommand{\bab}{\texttt{BaB}\xspace}
\newcommand{\mnist}{\texttt{MNIST}\xspace}
\newcommand{\ltwo}{\texttt{L2}\xspace}
\newcommand{\lfour}{\texttt{L4}\xspace}
\newcommand{\base}{\texttt{BASE}\xspace}
\newcommand{\deep}{\texttt{DEEP}\xspace}
\newcommand{\wide}{\texttt{WIDE}\xspace}

\newcommand{\cifar}{\texttt{CIFAR-10}\xspace}
\newcommand{\abcrown}{{$\alpha\beta$-\textsf{Crown}}\xspace}
\newcommand{\originNetwork}{\ensuremath{N}\xspace}
\newcommand{\ce}{\hat{\bm{x}}}
\newcommand{\true}{\ensuremath{\mathsf{true}}\xspace}
\newcommand{\false}{\ensuremath{\mathsf{false}}\xspace}
\newcommand{\verifier}{\ensuremath{\mathtt{AppVer}}\xspace}
\newcommand{\specDist}{\hat{p}}
\newcommand{\reluSpec}{\Gamma}
\newcommand{\reluAction}{r}
\newcommand{\reluActionPos}{r^+}
\newcommand{\reluActionNeg}{r^-}

\newcommand{\seman}[1]{\ensuremath{\llbracket#1\rrbracket}\xspace}
\newcommand{\specDistMin}{\hat{p}_{\mathrm{min}}}

\newcommand{\reluHeuristic}{\mathsf{H}}
\newcommand{\specTree}{\mathcal{T}}
\newcommand{\reward}{\ensuremath{\mathsf{R}}\xspace}

\newcommand{\myparagraph}[1]{\smallskip\noindent{\bf #1.}}
\newcommand{\tool}{\ensuremath{\mathsf{ABONN}}\xspace}

\newtheorem{mytheorem}{Theorem}
\newtheorem{mydefinition}[mytheorem]{Definition}{\bfseries}{\rmfamily}

\makeatletter
\newcommand{\linebreakand}{%
 \end{@IEEEauthorhalign}
 \hfill\mbox{}\par
 \mbox{}\hfill\begin{@IEEEauthorhalign}
}
\makeatother
  
\begin{document}

\title{
Adaptive Branch-and-Bound Tree Exploration for Neural Network Verification
}

\author{\IEEEauthorblockN{Kota Fukuda\IEEEauthorrefmark{1}, 
Guanqin Zhang\IEEEauthorrefmark{2}, 
Zhenya Zhang\IEEEauthorrefmark{1}, 
Yulei Sui\IEEEauthorrefmark{2},
Jianjun Zhao\IEEEauthorrefmark{1}
}

\IEEEauthorblockA{\IEEEauthorrefmark{1}Kyushu University, Fukuoka, Japan}
\IEEEauthorblockA{\IEEEauthorrefmark{2}University of New South Wales, Sydney, Australia}
}
\maketitle

\begin{abstract}
Formal verification is a rigorous approach that can provably ensure the quality of neural networks, and to date, Branch and Bound (BaB) is the state-of-the-art that performs verification by splitting the problem as needed and applying off-the-shelf verifiers to sub-problems for improved performance. However, existing BaB may not be efficient, due to its naive way of exploring the space of sub-problems that ignores the \emph{importance} of different sub-problems. To bridge this gap, we first introduce a notion of ``importance'' that reflects how likely a counterexample can be found with a sub-problem, and then we devise a novel verification approach, called \tool, that explores the sub-problem space of BaB adaptively, in a Monte-Carlo tree search (MCTS) style. The exploration is guided by the ``importance'' of different sub-problems, so it favors the sub-problems that are more likely to find counterexamples. As soon as it finds a counterexample, it can immediately terminate; even though it cannot find, after visiting all the sub-problems, it can still manage to verify the problem. We evaluate \tool with 552 verification problems from commonly-used datasets and neural network models, and compare it with the state-of-the-art verifiers as baseline approaches. Experimental evaluation shows that \tool demonstrates speedups of up to $15.2\times$ on MNIST and $24.7\times$ on CIFAR-10. We further study the influences of hyperparameters to the performance of \tool, and the effectiveness of our adaptive tree exploration.
\end{abstract}


\section{Introduction}
Recently, artificial intelligence (AI) has experienced an explosive development and pushes forward the state-of-the-art in various domains. Due to their advantages in handling complex data (e.g., images and natural languages), AI products, especially \emph{deep neural networks}, have been deployed in different safety-critical systems, such as autonomous driving and medical devices. Despite such prosperity,  a surge of concerns also arise about their safety, because misbehavior of those systems can pose severe threats to human society~\cite{liu2021algorithms}. Given that neural networks are notoriously vulnerable to adversarial perturbations~\cite{goodfellow2015explaining}, it is of great significance to ensure their quality before their deployment in real world.



As a rigorous approach, formal verification has been actively studied in recent years~\cite{liu2021algorithms,muller2022third}, which can 
provably ensure the quality of neural networks. 
A straightforward approach~\cite{cheng2017maximum, tjeng2018evaluating}  is by encoding a verification problem as logical constraints which can be solved by mixed integer linear programming (MILP). However, due to the non-linearity of activation functions, this approach  is not scalable to large networks.
Abstraction-based approaches~\cite{singh2018fast,singh2019abstract,zhang2018efficient,wong2018provable}, that over-approximate the output regions of networks, are much faster; however, they suffer from an \emph{incompleteness issue} and may often raise \emph{false alarms} that are misleading, with spurious counterexamples.

Branch and Bound (\bab)~\cite{bunel2020branch} is an advanced verification approach for neural networks that can mitigate the incompleteness issue of approximated verifiers. It is essentially a ``\emph{divide-and-conquer}'' strategy, which splits a problem if false alarms arise and applies approximated verifiers to the sub-problems for which those verifiers suffer less from the incompleteness issue. After verifying all the sub-problems or encountering a real counterexample in a sub-problem, \bab can conclude the verification and return accordingly. In this way, it overcomes the incompleteness issue in directly applying approximated verifiers to the original verification problem, thus more effective.


\myparagraph{Motivation} While \bab has shown great promise, it may not be efficient, because it explores the space of sub-problems naively in a ``\emph{breadth-first}'' manner and ignores the different ``\emph{importance}'' of different sub-problems. Indeed, different sub-problems should have different priorities in verification, because with some sub-problems, it is more likely to find counterexamples, and thereby conclude the verification efficiently. 
By prioritizing these more important sub-problems, we can save plenty of efforts in checking unnecessary sub-problems and improve the efficiency of verification.

\myparagraph{Contributions}
In this paper, we propose a novel verification approach \tool, that is, \underline{\textbf{A}}daptive \underline{\textbf{B}}aB with \underline{\textbf{O}}rder for \underline{\textbf{N}}eural \underline{\textbf{N}}etwork verification, which incorporates the insights above.

We first introduce a notion of ``\emph{importance}'' over different sub-problems, called \emph{counterexample potentiality}, used to characterize the likelihood of finding counterexamples with a sub-problem. It is defined based on two attributes of a sub-problem, including its fineness (i.e., the level of problem splitting) and a quantity returned by approximated verifiers that signify the level of specification violation of  the sub-problem. 

We then devise \tool, which features an adaptive  exploration in the space of  sub-problems in a \emph{Monte-Carlo
tree search (MCTS)}~\cite{browne2012survey} style. Guided by the counterexample potentiality of different sub-problems,
\tool favors the sub-problems that are more likely to find counterexamples, and as soon as it finds a counterexample, it can immediately terminate and report a negative result about the verification problem. Even though it cannot find such a counterexample, after visiting
all the sub-problems, it can still manage to verify the problem. 

 We perform a comprehensive experimental evaluation for \tool, with 552 verification problems that span over five neural network models in datasets MNIST and CIFAR-10, and two state-of-the-art verification approaches as baselines. Our evaluation  shows a great speedup of \tool over the baseline approaches, for up to 15.2$\times$ in
MNIST, and up to 24.7$\times$ in CIFAR-10. We further study the 
influences of hyperparameters, thereby exhibiting the necessity of striking a balance between ``exploration'' and ``exploitation'' in MCTS. Moreover, we demonstrate the effectiveness of our adaptive tree exploration strategy by separating the results for violated problems and certified problems.

\section{Overview of The Proposed Approach}

\subsection{Verification Problem and \bab Approach}

\begin{figure}[!tb]
\centering
    \begin{subfigure}[b]{0.4\textwidth}
        \centering
        \includegraphics[width=\textwidth]{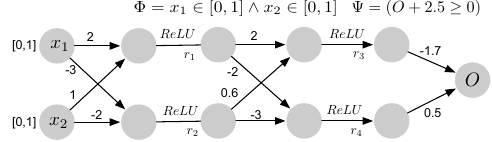}
    \caption{A neural network $\originNetwork$ and its specification}
    \label{fig:network}
    \end{subfigure}
    \begin{subfigure}[b]{0.4\textwidth}
    \centering
\includegraphics[width=\textwidth]{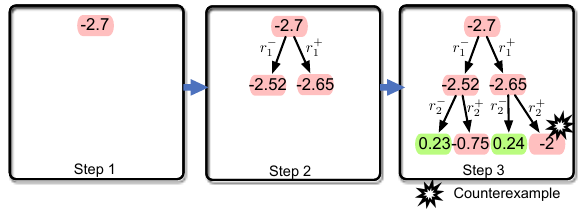}
    \caption{A running example of \bab for verifying $\originNetwork$}
    \label{fig:babrun}
    \end{subfigure}
    \caption{Neural network verification problem and a solution via branch and bound (\bab).}\label{fig:exa}
\end{figure}

A neural network $\originNetwork$ and its associated specification are given in Fig.~\ref{fig:exa}. The verification problem aims to determine whether the neural network $\originNetwork$ satisfies the specification, i.e., for all inputs $(x_1, x_2)\in [0,1]\times [0,1]$, whether it holds that the output $O$ of $\originNetwork$ satisfies a logical constraint $O+2.5 \ge 0$.

Fig.~\ref{fig:babrun} illustrates how \bab~\cite{bunel2020branch} works to solve the problem.
\bab splits the problem and applies verifiers to sub-problems to pursue better performance.
It decides whether a (sub-)problem should be split, by applying an approximated verifier to it. The verifier can return a value, as depicted in each node in Fig.~\ref{fig:babrun}, which signifies \emph{how far} each problem is from being violated. If the value is positive, then the problem is verified and no need to split.
Otherwise, \bab will check whether the negative value returned by the verifier is a false alarm, by validating the \emph{counterexample} given by the verifier.  
If the counterexample is a real one, the verification can be concluded that the specification is violated. Otherwise, \bab will split the problem and apply verifiers to sub-problems. Problem splitting can be done in different ways; in this paper, we follow existing literature~\cite{bunel2020branch,ugare2023incremental} and impose input constraints to ReLU activation functions.

The  running example in Fig.~\ref{fig:babrun} works as follows:
\begin{compactenum}[1)]
    \item By applying a verifier to the original problem, \bab obtains a negative returned value $-2.7$, and identifies that this is a false alarm. So, it splits the problem to two sub-problems;
    
    \item\label{step:secondStep} \bab then applies the verifier to each of the two sub-problems, by which it obtains $-2.52$ and $-2.65$, respectively. Again, it identifies that both are false alarms, and so, it continues to split the sub-problems;
    
    \item This continues in the subsequent layer of the tree, and it identifies the sub-problems that are verified (with values $0.23$ and $0.24$) and that need to be split further (with values $-0.75$). Notably, the verification can be terminated in this layer, because a real counterexample is detected in the sub-problem with $-2$, which can serve as an evidence that the network does not satisfy the specification. 
\end{compactenum}

\begin{figure}[!t]
    \centering
    \includegraphics[width=0.8\linewidth]{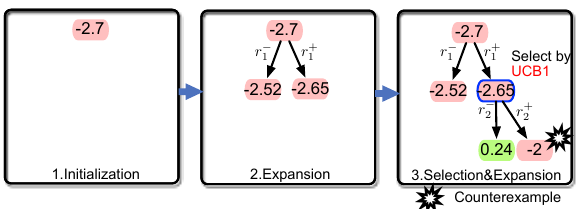}
    \caption{\tool's process for solving the problem.}
    \label{fig:Oliverun}
\end{figure}

\subsection{The Proposed Approach}
While \bab in Fig.~\ref{fig:babrun} is effective to solve the problem, it may be not efficient, because it ignores the information about \emph{how important} each sub-problem is. Indeed, different sub-problems are not equivalent in terms of their importance, and in particular, the exploration of the \bab tree should favor the sub-problems that are more likely to find a counterexample, because once that is achieved, verification can be early terminated. 

Moreover, the information about the likelihood of finding counterexample has been embedded in the attributes of the nodes in the \bab tree. For instance, among the children of the root, the node with $-2.65$ should be prioritized than the node with $-2.52$, because according to such an assessment done by approximated verifiers, the former involves a sub-problem that is farther from being verified than the latter, thus more likely to contain a real counterexample.

Having these insights, we propose \tool, 
that explores the space of sub-problems adaptively, in a  \emph{Monte-Carlo tree search (MCTS)} fashion. MCTS~\cite{browne2012survey} is a \emph{reinforcement learning}-based algorithm that expands a search tree guided by \emph{rewards}, and so, it pays more efforts to the branches that are more ``\emph{promising}''; meanwhile, to avoid being too greedy, it also allocates budgets to the less ``\emph{promising}'' branches to strike a balance.

For the example in Fig.~\ref{fig:exa}, we differ from the naive \bab in the selection of branches. As shown in Fig.~\ref{fig:Oliverun}, \tool selects the most ``\emph{promising}'' node over the children of the root to proceed, and manages to find a real counterexample in the subsequent steps. Notably, \tool is more efficient than the naive \bab, because it saves two visits of the sub-problems, each of which involves an expensive process of problem solving. 




\section{Preliminaries}\label{sec:preliminary}
\myparagraph{Neural Networks} 
A (feed-forward) neural network $\originNetwork: \mathbb{R}^n \to \mathbb{R}^m$ (see the example in Fig.~\ref{fig:network}) maps an $n$-dim input to an $m$-dim output. 
The mapping from the input to output is conducted by alternating between affine transformations $\hat{\bm{x}}_i = W_i\bm{x}_{i-1} + B_i$ (where $W_i$ is a matrix of \emph{weights} and $B_i$ is a vector of \emph{biases}) and non-linear transformations $\bm{x}_i = \sigma(\hat{\bm{x}}_i)$ (where $\sigma$ is called an \emph{activation function}). 
Specially, $\bm{x}_0$ is the input of $\originNetwork$ and $\bm{x}_L$ is the output of $\originNetwork$, where $L$ is the number of \emph{layers} of the network. Regarding the selection of activation function, we follow many existing works (see a survey~\cite{liu2021algorithms}), and adopt ReLU (i.e., $\sigma(x) = \max(0, x)$) as the activation function.

\myparagraph{Verification Problem} 
We denote the \emph{specification} of a neural network $\originNetwork$ as a tuple $(\Phi, \Psi)$, where $\Phi$ is a predicate over the input of $\originNetwork$, and $\Psi$ is a predicate over the output of $\originNetwork$. Commonly-used properties, such as \emph{local robustness} in image classification, can be described by such a specification. Given a reference input $\bm{x}_0$, $\Phi(\bm{x})$ requires that the input $\bm{x}$ must stay in the region $\{\bm{x} \mid \|\bm{x} - \bm{x}_0\|_\infty \leq \epsilon \}$, where $\epsilon\in \mathbb{R}$ denotes a perturbation distance, and $\Psi(\originNetwork(\bm{x}))$ requires that the output $\originNetwork(\bm{x})$ must imply the same label as that of $\bm{x}_0$.

A verification problem concerns with a question as follows: 
\begin{compactitem}
    \item \textit{Given:} a neural network $\originNetwork$ and a specification $(\Phi, \Psi)$;
    \item \textit{Return:} \true if $\Psi(\originNetwork(\bm{x}))$ holds, for any input $\bm{x}$ that holds $\Phi(\bm{x})$; \false otherwise.
\end{compactitem}
A \emph{verifier} is implemented to answer a verification problem. In case it returns \false, it will simultaneously return a \emph{counterexample} $\ce$, which is an input that holds $\Phi(\ce)$ but does not hold $\Psi(\ce)$, as an evidence of the violation of specification.

\myparagraph{Branch and Bound (\bab)}
\bab~\cite{bunel2020branch} is the state-of-the-art neural network verification approach, and has been adopted as the theoretical foundation of several verification tools, such as \abcrown~\cite{wang2021beta}. While it is essentially a ``divide-and-conquer'' strategy, now it is often used to mitigate the \emph{incompleteness} issue of approximated verifiers, by applying these verifiers to smaller verification problems. Below, we elaborate on how \bab collaboratively works with approximated verifiers. 


Approximated verifiers, denoted as \verifier, are a class of verifiers that solve a problem by over-approximating the output region of a neural network---once the over-approximated output satisfies the specification, the original output must also satisfy it. By applying to a problem, \verifier can return a real value $\specDist$ to indicate the satisfaction of the over-approximated output: if $\specDist$ is positive, then the over-approximated output satisfies the specification; otherwise, it violates the specification. However, in the latter case, the result given by \verifier is not necessarily correct, because it is possible that the original output does not violate the specification. This can be validated by checking the counterexample $\hat{\bm{x}}$ returned by \verifier, and if $\Psi(\originNetwork(\hat{\bm{x}}))$ is not violated, then $\hat{\bm{x}}$ is a spurious one, and so \verifier raises a false alarm. This is known as the \emph{incompleteness issue} of \verifier.



\bab is used to mitigate this issue, by applying \verifier adaptively to smaller problems. It works as follows:
\begin{compactenum}[i)]
    \item First, it applies \verifier to the original verification problem, and receives $\specDist$: if $\specDist$ is positive, or $\specDist$ is negative with a valid counterexample $\ce$, verification can be terminated; 
    \item \label{step:case} In the case that $\specDist$ is negative and the counterexample $\ce$ is a spurious one, it splits the problem into two sub-problems. This can be done by adding an additional logical constraint to each of the sub-problems, which predicates over the input condition of the ReLU  in a neuron selected from the network (i.e., the additional constraint is either the input of ReLU is positive, or the input of ReLU is negative);
    \item It then applies \verifier to each of the sub-problems, and decides whether it needs to further split the sub-problems, in the same way as that in Step~\ref{step:case}.
\end{compactenum}


\myparagraph{\bab Tree} The process of problem splitting and solving in \bab can be characterized as a tree, as illustrated in Fig.~\ref{fig:babrun}. In this tree, each node identifies a sub-problem, and each sub-problem is identified by a sequence $\reluSpec$ of input constraints of ReLUs (namely, the ReLU constraints from the root to the current node). Given the $i$-th ReLU in a network, let $\reluActionPos_i$ denote the condition that the input of the ReLU is positive, and $\reluActionNeg_i$ denote the condition that the input of the ReLU is negative. Specially, the root is identified by an empty sequence. 


It remains a problem in \bab how  a ReLU should be selected in the network to split a problem, when a false alarm arises with the problem. There have been various strategies devised for that purpose, such as DeepSplit \cite{henriksen2021deepsplit} and FSB~\cite{de2021improved}. All these strategies can be seen as a pre-defined heuristic $\reluHeuristic$ that can return a ReLU given a specific (sub-)problem (namely,  given a sequence of ReLUs that have already been expanded). In this work, our aim is not to optimize $\reluHeuristic$, but we are orthogonal to that line of work (see~\S{}\ref{sec:related}); we simply select the state-of-the-art ReLU selection method~\cite{henriksen2021deepsplit},  following existing literature \cite{ugare2023incremental}.


\section{\tool: The Proposed Verification Approach}\label{sec:proposeidea}
As introduced, \bab produces a large space of sub-problems and accomplishes verification by exhaustively visiting the sub-problems. However, the strategy adopted by existing \bab for space exploration is naive, in the sense that it ignores the importance of different sub-problems, but visits them in a naive ``\emph{first come, first serve}'' manner, which is very inefficient.

In this paper, we leverage a notion of ``importance'' over the sub-problems (detailed in~\S{}\ref{sec:cepo}) and devise a novel \bab-based verification approach that features an adaptive tree exploration in a \emph{Monte-Carlo tree search (MCTS)} style (detailed in~\S{}\ref{sec:mctsApproach}). Notably, we link the notion of ``importance'' to the \emph{likelihood} of finding counterexamples in different sub-problems, and the rationale behind is as follows: during verification, we prioritize those sub-problems that are more likely to find counterexamples; as soon as we find a counterexample, we can immediately terminate the verification and return \false as a result; even though we cannot find such a counterexample, compared to naive \bab, we only differ in the order we visit the sub-problems but we can still manage to verify the problem finally.


\subsection{Counterexample Potentiality}\label{sec:cepo}
We elaborate on the notion of ``importance'' of different sub-problems, by defining an order called \emph{counterexample potentiality}. Specifically, the potentiality of having a counterexample in a node of \bab tree is related to the following two attributes:
\begin{compactitem}[$\bullet$]
\item \emph{Node depth.}
In \bab tree, the more a problem is split, the less over-approximation will be introduced when applying \verifier; therefore, under the premise that $\specDist$ is still negative for a node $\reluSpec$, the greater its depth is, the more likely it is to find a real counterexample from $\reluSpec$;
\item \emph{$\specDist$.} The value $\specDist$, obtained by applying \verifier to a sub-problem, is a quantity that reflects \emph{how far} the sub-problem is from being violated, according to the evaluation of \verifier that relies on the over-approximation of the network output. Although it is not precise, under a fixed \verifier, this value is correlated with the real satisfaction level that can be computed by the real output region, and so it can be used to indicate the potentiality of counterexamples. Specifically, in the case where $\specDist$ is negative, the greater $|\specDist|$ is, the more likely that the sub-problem contains a counterexample. 
\end{compactitem}
Based on the two node attributes mentioned above, we define  counterexample potentiality of a node in Def.~\ref{def:cepo}.

\smallskip
\begin{mydefinition}[Counterexample potentiality]\label{def:cepo}
Let $\reluSpec$ be a node that has depth $\mathtt{depth(\reluSpec)}$ and verifier evaluation $\specDist$ (with a counterexample $\ce$ if $\specDist < 0$). The counterexample potentiality $\seman{\reluSpec}\in [0,1]\cup \{+\infty, -\infty\}$ of $\reluSpec$ 
is computed as follows:
\begin{align*}
    \seman{\reluSpec} := \begin{cases}
        -\infty  & \text{if } \specDist > 0 \\
        +\infty  & \text{if } \specDist<0 \text{ and } \mathtt{valid}(\ce) \\
        \lambda\frac{\mathtt{depth}(\reluSpec)}{K} + (1-\lambda)\frac{\specDist}{\specDistMin} & \text{otherwise}
    \end{cases}
\end{align*}
where $\lambda\in [0,1]$ is a parameter that controls the weights of the two attributes, and $K$ is the total number of neurons (i.e., ReLUs) in the network.
\end{mydefinition}

Def.~\ref{def:cepo} characterizes the likelihood of containing counterexamples in a node $\reluSpec$, in the following way:
\begin{compactitem}[$\bullet$]
    \item If $\specDist > 0$, there is no chance to find a counterexample in $\reluSpec$, and so $\seman{\reluSpec} = -\infty$;
    \item If $\specDist < 0$ and $\ce$ is a valid one, it means that a counterexample is already found, so $\seman{\reluSpec} = +\infty$;
    \item Otherwise, $\seman{\reluSpec}$ is determined by the node attributes as mentioned above, and we use a hyperparameter $\lambda$ to adjust the weights of the two attributes.
\end{compactitem}

In~\S{}\ref{sec:mctsApproach}, we present our efficient \bab-based verification approach, by exploiting this counterexample potentiality to guide the exploration of sub-problem space.



\subsection{MCTS-Style Tree Exploration for \bab-Based Verification}\label{sec:mctsApproach}

\begin{algorithm}[!tb]
\caption{\tool: An MCTS-style verification algorithm}
\label{alg:mcts_nn_verify}
\footnotesize
\begin{algorithmic}[1]
\Require A neural network $\originNetwork$, input and output specification $\Phi$ and $\Psi$, an approximated verifier $\verifier(\cdot)$, a ReLU selection heuristic $\reluHeuristic(\cdot)$, and hyperparameters $\lambda$ and $c$.
\Ensure A $\mathit{verdict} \in \{\true, \false, \mathtt{timeout}\}$ 

\Statex
\State $\langle\specDist, \ce\rangle\gets \verifier(\originNetwork, \Phi, \Psi, \varepsilon)$ \label{line:appToOrigin}
\State $\reward(\varepsilon)\gets \seman{\varepsilon}$ \label{line:originR}
\State $\specTree(\varepsilon) \gets \{\varepsilon\}$ \label{line:originTree}
\If{$\specDist<0$ and $\mathtt{not\;valid}(\ce)$}
\While{not reach termination condition}\label{line:whileLoop}
\State \Call{MCTS-BaB}{$\varepsilon, \originNetwork, \Phi, \Psi$}\label{line:callGreedyBaBRoot}
\EndWhile
\State\label{line:return}\Return
$
\begin{cases}
    \true & \text{if } \reward(\varepsilon) = -\infty \\
    \false & \text{if } \reward(\varepsilon) = +\infty \\
    \mathtt{timeout} & \text{otherwise}
\end{cases}
$
\Else \label{line:originConclude}
\State \label{line:originReturn}\Return
$
\begin{cases}
    \true & \text{if }\specDist > 0 \\
    \false & \text{if } \specDist < 0 \text{ and } \mathtt{valid}(\ce)
\end{cases}
$
\EndIf
\Statex
\Function{MCTS-BaB}{$\reluSpec, \originNetwork, \Phi, \Psi$}
\State $\reluAction_k\gets \reluHeuristic(\reluSpec)$ \label{line:reluSelect} \Comment{select a ReLU}
\If{$\reluSpec \cdot \reluActionPos_k \in \specTree$}\label{line:checkExists}
\State $\reluSpec^* \gets \argmax_{a\in\{\reluActionPos_k, \reluActionNeg_k\}} \left( \reward(\reluSpec\cdot a) + c\sqrt{\frac{2\ln |\specTree(\reluSpec)|}{|\specTree(\reluSpec\cdot a)|}} \right)$\label{line:UCB1}
\vspace{2pt}
\Statex\Comment{Select a child by UCB1}
\State \Call{MCTS-BaB}{$\reluSpec^*, \originNetwork, \Phi, \Psi$}\label{line:recursiveCall} \Comment{recursive call}
\Else
\For{$a\in\{\reluActionPos_k, \reluActionNeg_k\}$}\label{line:expandChildren}
    \State $\langle\specDist, \ce\rangle \gets \verifier(\originNetwork, \Phi, \Psi, \reluSpec \cdot a)$ \label{line:verifierSubTree}
    \vspace{2pt}
    \State $\reward(\reluSpec\cdot a) \gets    \seman{\reluSpec\cdot a}$ \label{line:rewardSubtree}
    \State $\specTree(\reluSpec\cdot a) \gets \{\reluSpec\cdot a\}$ \label{line:updateTree}
\EndFor
\EndIf
\State $\reward(\reluSpec)\gets \argmax_{a\in\{\reluActionPos_k, \reluActionNeg_k \}}{\seman{\reluSpec\cdot a}}$\label{line:backpropagateR} \Comment{back-propagate rewards}
\vspace{2pt}
\State $\specTree(\reluSpec) \gets \specTree(\reluSpec) \cup \specTree(\reluSpec \cdot \reluActionPos_k) \cup \specTree(\reluSpec \cdot \reluActionNeg_k)$ \label{line:backpropagateTree} \Comment{record new nodes}
\EndFunction
\end{algorithmic}
\end{algorithm}

We present our MCTS-style \bab tree exploration algorithm for verification of neural networks. The idea is that, we use counterexample potentiality to guide the search towards the sub-problems that are more likely to find counterexamples, so compared to the naive \bab, our tree exploration is imbalanced and favors the branches that are more ``\emph{promising}''. 


\smallskip
Our verification algorithm is presented in Alg.~\ref{alg:mcts_nn_verify}, and see Fig.~\ref{fig:Oliverun} for an    illustration of the different stages of the algorithm. 

\myparagraph{Initialization}
Alg.~\ref{alg:mcts_nn_verify} starts with handling the root node, which identifies the original verification problem. It applies \verifier to the problem, and obtains the returned $\specDist$ and $\ce$ (in case $\specDist < 0$) (Line~\ref{line:appToOrigin}). If $\specDist > 0$, or $\specDist < 0$ and $\ce$ is valid, the verification can be concluded (Line~\ref{line:originConclude}); otherwise, i.e., $\specDist < 0$ is a false alarm, so the problem needs to be split. Here, we use $\reward(\reluSpec)$ to record the \emph{reward} of the node $\reluSpec$ (Line~\ref{line:originR}), and $\specTree(\reluSpec)$ to record the (sub-)tree (i.e., the set of nodes in the tree) that has $\reluSpec$ as its root (Line~\ref{line:originTree}). Later, we will show that these two structures are helpful to explore the tree by the mechanism of MCTS.

\myparagraph{Expansion}
The function \textsc{MCTS-BaB} goes through an MCTS workflow. Given a node $\reluSpec$, it first checks whether the children of $\reluSpec$ have been expanded.  If not yet, it expands the children of $\reluSpec$, by splitting the problem identified by $\reluSpec$ into sub-problems. The two sub-problems are respectively identified by $\reluSpec\cdot \reluActionPos_k$ and $\reluSpec\cdot \reluActionNeg_k$, where $\reluAction_k$ is the ReLU selected by $\reluHeuristic$ in Line~\ref{line:reluSelect}. The algorithm applies \verifier to each of the sub-problems (Line~\ref{line:verifierSubTree}). Based on the results, it computes the counterexample potentiality and records it  as \emph{reward} (Line~\ref{line:rewardSubtree}), and uses  $\specTree$ to record the newly expanded children (Line~\ref{line:updateTree}).

\myparagraph{Back-Propagation} After expansion, it propagates the rewards and visits backwards to the ancestors of the new nodes, until the root node. In terms of rewards, the reward of the parent node will be updated to be the maximal reward of its children (Line~\ref{line:backpropagateR}); moreover, the newly expanded nodes will be added to $\specTree$ of the parent node (Line~\ref{line:backpropagateTree}).

\myparagraph{Selection} If all children of a node $\reluSpec$ have been expanded, it needs to select a child to proceed. In line with normal MCTS, it selects by UCB1~\cite{browne2012survey}, which is an algorithm that favors not only the branches that have greater rewards, but also considers the branches that are less visited, because rewards may not always be accurate to reflect the likelihood of counterexample existence in different sub-problems. UCB1 algorithm is given in Line~\ref{line:UCB1} of Alg.~\ref{alg:mcts_nn_verify}: it selects a child of $\reluSpec$ by comparing an  aggregated value of their rewards and visits. In particular, the first term is the reward of each child, and the second term is inversely proportional to the number of visits of each child; a hyperparameter $c$ is used to strike a balance of the two terms.

\myparagraph{Termination} The algorithm is terminated until the termination condition is reached (Line~\ref{line:whileLoop}). In particular, there are three conditions, and it suffices to meet one of them to terminate:
\begin{compactitem}[$\bullet$]
    \item $\reward(\epsilon) = +\infty$: it implies that a real counterexample has been found, and so verification can be terminated with \false;
    \item $\reward(\epsilon) = -\infty$: it implies that all of sub-problems have been verified, and so verification can be terminated with \true;
    \item \emph{timeout}: If the time budget is used up, verification should also be terminated without meaningful conclusion.
\end{compactitem}

\section{Experimental Evaluation} 
\subsection{Experiment Settings}
\myparagraph{Baseline and Metrics} We compare with two state-of-the-art verification approaches, namely, \bab-baseline and \abcrown. The evaluation metrics include the number of instances solved and the average time cost of each approach. We also compute the speedup of \tool for individual verification problems w.r.t. the \bab-baseline. Our baselines are as follows:
\begin{compactitem}[$\bullet$]
    \item \bab-baseline: The naive \bab as introduced in~\S{}\ref{sec:preliminary}, i.e., it explores sub-problem space in a ``breadth-first'' manner;
    
    \item \abcrown \cite{zhang2018efficient,wang2021beta}: 
    The state-of-the-art verification tool according to~\cite{muller2022third}, that features various sophisticated heuristics for performance improvement.
\end{compactitem}
Our implementation of \tool adopts the same approximated verifiers~\cite{singh2019abstract, singh2018deepz} and ReLU selection heuristic~\cite{henriksen2021deepsplit}  as that of \bab-baseline. For the hyperparameters in Alg.~\ref{alg:mcts_nn_verify}, as a default version of our tool, we set $\lambda$ as 0.5 and $c$ as 0.2; we will study the influences of these two hyperparameters in RQ2. All the code and data are available online\footnote{\url{https://github.com/DeepLearningVerification/ABONN}}.

\begin{wrapfigure}[10]{r}{0.5\linewidth}
\vspace{-10pt}
\includegraphics[width=\linewidth]{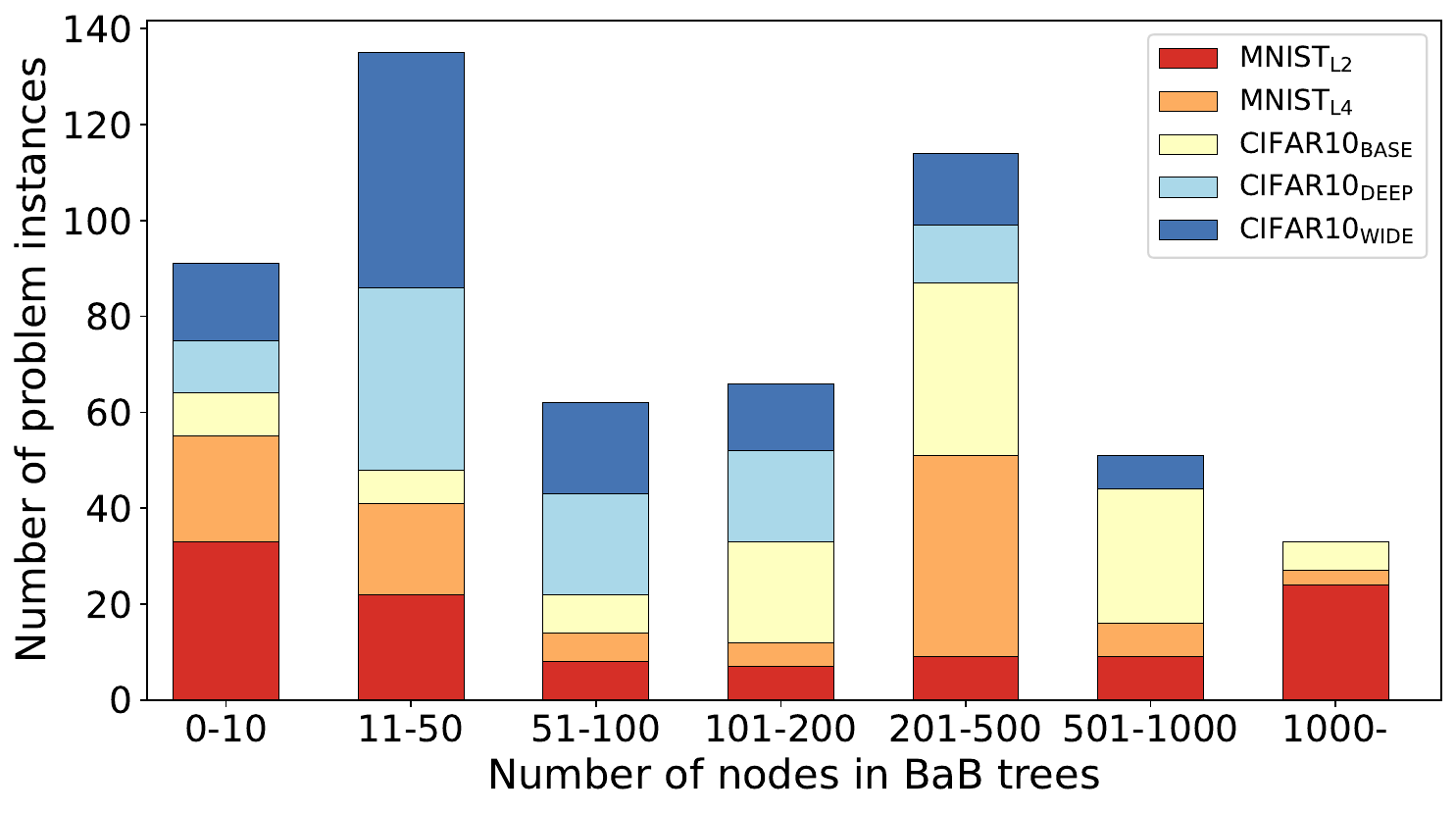}
\caption{The distribution of the sizes of the \bab trees used in our experiments}\label{fig:dis_boxplot}
\end{wrapfigure}
\myparagraph{Benchmarks}
We adopt 552 verification problems about $L_\infty$-based local robustness for \mnist and \cifar. These datasets are standard benchmarks
in community, widely recognized in VNN-COMP~\cite{muller2022third}, an annual competition of neural network verification. We select meaningful problems that are neither too easy nor too hard to solve, as evidenced by the distribution of the tree sizes with \bab-baseline in Fig.~\ref{fig:dis_boxplot}. 
We evaluate two fully connected networks with MNIST and three convolutional networks with CIFAR-10, as shown in Table~\ref{tab:benchmarks}. Moreover, Table~\ref{tab:benchmarks} also presents the number of specifications considered for each model. 

\begin{table}[!tb]
    \centering
    \caption{Details of the benchmarks}\label{tab:benchmarks}
    \scalebox{0.80}{
    \begin{tabular}{c|c|c|c|c}
    \toprule
        Model Dataset & Architecture & Dataset & \#Neurons &\# Instances\\\hline
        $\mnist_{{{\ltwo}}}$ &  2 $\times$ 256 linear  & \mnist & 512 & 112\\ 
        $\mnist_{{{\lfour}}}$ &  4 $\times$ 256 linear  & \mnist & 1024 &104\\\hline 
        $\cifar{{\base}}$  & 2 Conv, 2 linear  & \cifar & 4852 & 115\\  
        $\cifar{{\wide}}$  & 2 Conv, 2 linear  & \cifar & 6244 & 101\\ 
        $\cifar{{\deep}}$  & 4 Conv, 2 linear & \cifar & 6756 & 120\\ \bottomrule
    \end{tabular}}
    \label{tab:models}
\end{table}


\myparagraph{Experiment Environment} 
Experiments ran on an AWS EC2 instance (8-core Xeon E5 2.90GHz, 16GB RAM) with 1000s timeout per problem. GUROBI 9.1.2 was used as the solver.


\subsection{Evaluation Results}
\begin{table}[!tb]
    \centering
    \small
        \caption{RQ1 -- Overall comparison of time consumption (in seconds) and the total number of solved instances.}
    \label{tab:comparisonWithBaB}
\scalebox{0.75
}{

    \begin{tabular}{l|rc|rc|rc}
    \toprule
        Model Dataset &  \multicolumn{2}{c}{\bab-baseline}  & \multicolumn{2}{c}{\abcrown} & 
        \multicolumn{2}{c}{\tool} \\ 
        
               & Solved  &Time  & Solved & Time  & Solved &Time  \\\hline
$\mnist_{{{\ltwo}}}$ & 95& 245.11& 96& 19.53& 92& 248.29 \\
$\mnist_{{{\lfour}}}$ & 59& 200.68& 43& 360.97& 57& 270.48 \\
$\cifar_{{\base}}$ & 27& 782.31& 32& 699.77& \tbgreen 106& \tbgreen 176.87 \\
$\cifar_{{\deep}}$ & 23& 749.74& 40& 516.25& \tbgreen 67& \tbgreen 369.58 \\
$\cifar_{{\wide}}$ & 26& 706.04& 38& 520.3&\tbgreen 75& \tbgreen 246.03 \\
\hline
        \bottomrule
    \end{tabular}}

\end{table}

\begin{figure*}
    \centering
    \begin{subfigure}[b]{0.19\linewidth}
        \includegraphics[width=\textwidth]{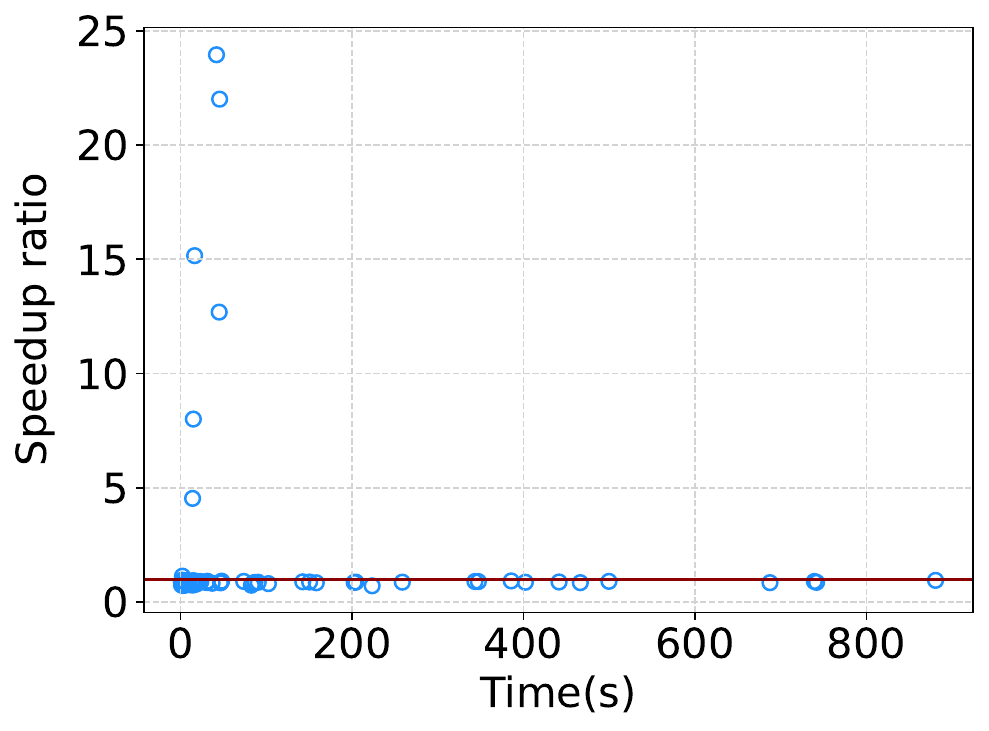}
        \caption{$\mnist_{\ltwo}$}
    \end{subfigure}
    \begin{subfigure}[b]{0.19\linewidth}
        \includegraphics[width=\textwidth]{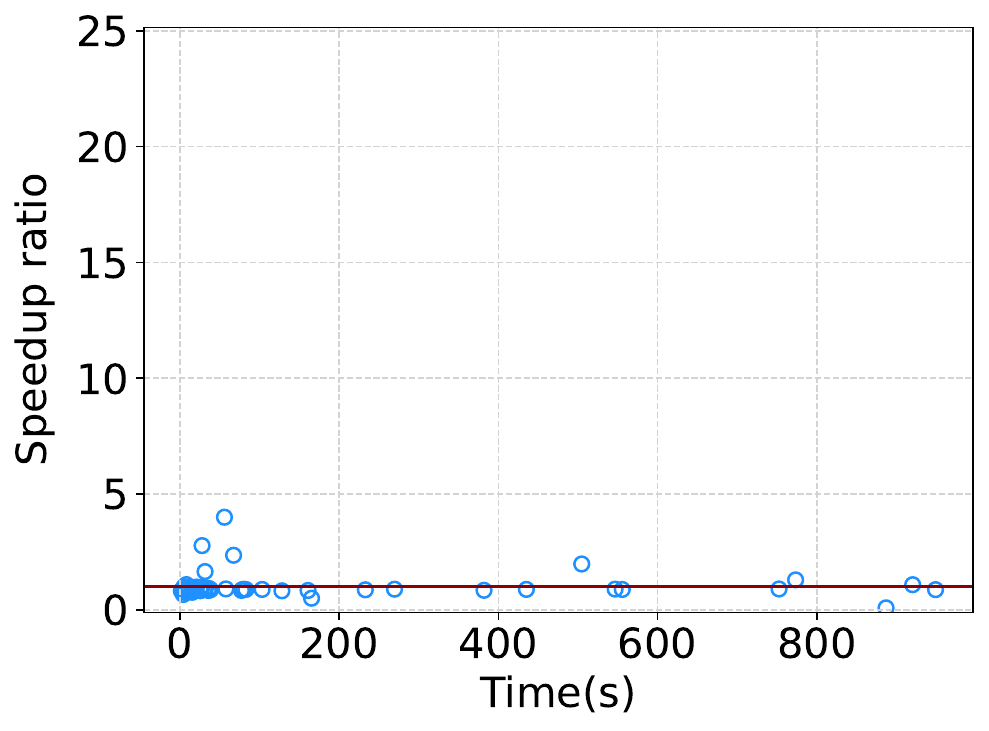}
        \caption{$\mnist_{\lfour}$}
    \end{subfigure}
    \begin{subfigure}[b]{0.19\linewidth}
        \includegraphics[width=\textwidth]{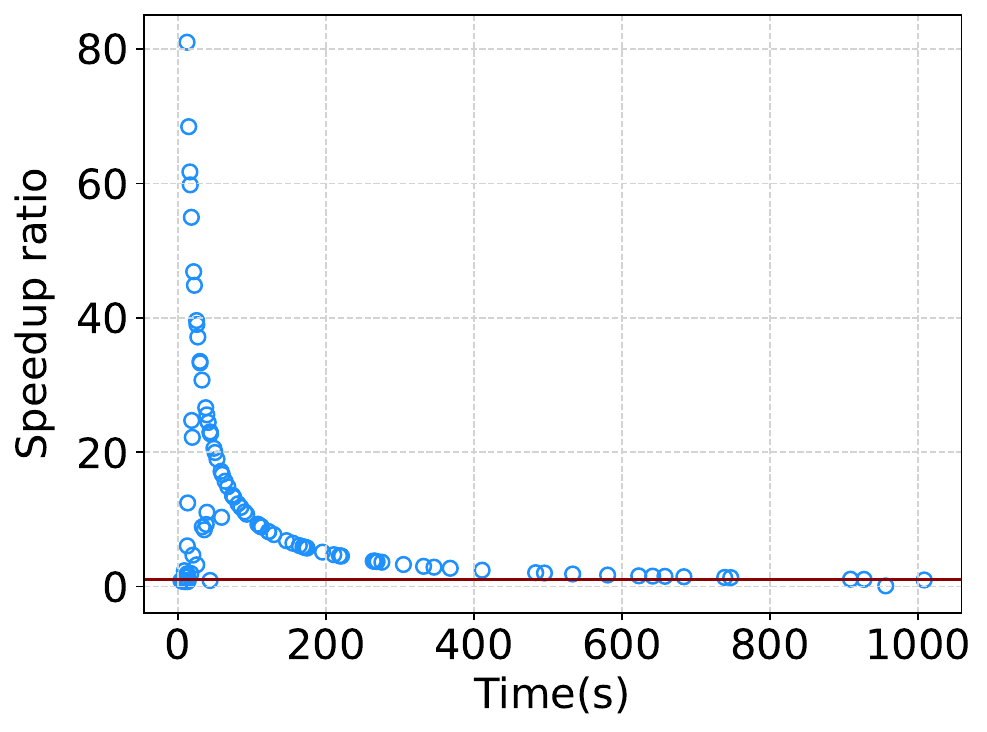}
        \caption{$\cifar_{{\base}}$}
    \end{subfigure}
    \begin{subfigure}[b]{0.19\linewidth}
        \includegraphics[width=\textwidth]{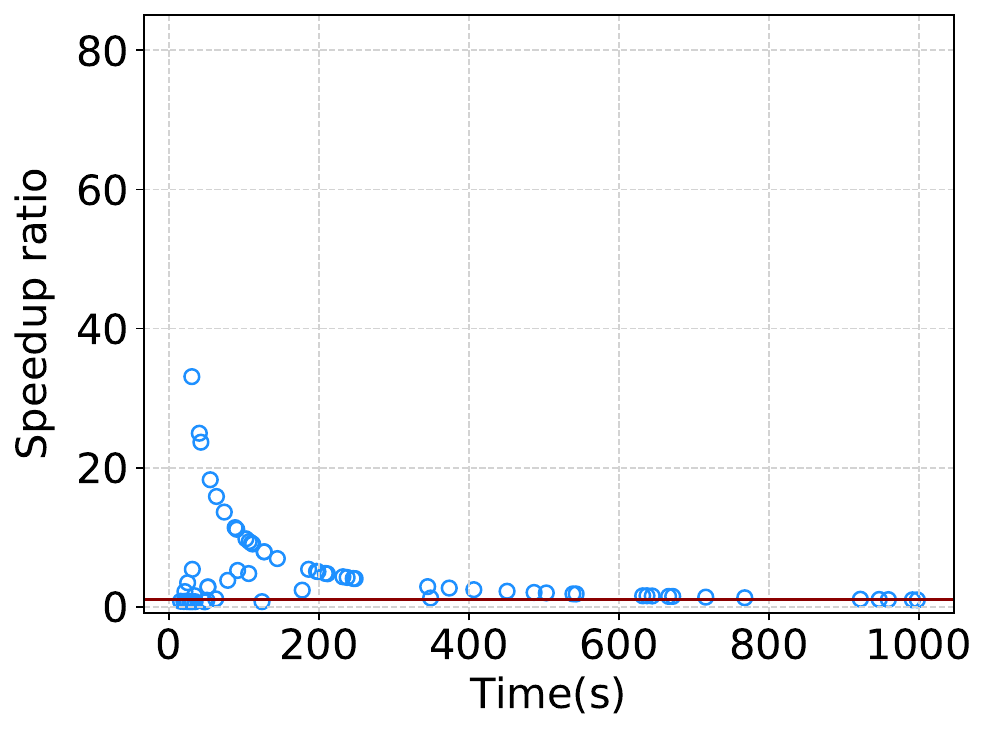}
        \caption{$\cifar_{{\deep}}$}
    \end{subfigure}
    \begin{subfigure}[b]{0.19\linewidth}
        \includegraphics[width=\textwidth]{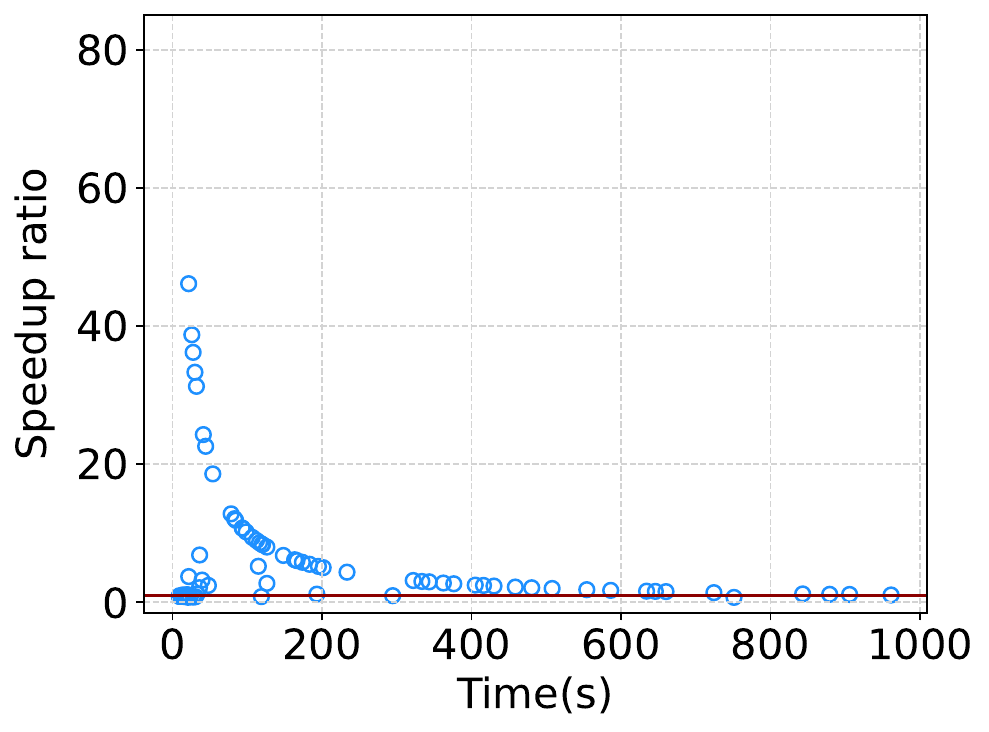}
        \caption{$\cifar_{{\wide}}$}
    \end{subfigure}
    \caption{RQ1 -- Comparison of \tool over \bab-baseline in time costs and speedup. Each blue dot stands for a  problem.}
    \label{fig:speedup}
\end{figure*}

\myparagraph{RQ1: Efficiency of \tool compared to baselines}

Table~\ref{tab:comparisonWithBaB} shows the number of solved problems across \mnist and \cifar and the average time costs of each approach. \tool demonstrates significant advantages in the \cifar models that are relatively complex. For instance, in $\cifar_{{\base}}$, it solves 79 more instances than \bab-baseline, and 74 more than \abcrown within the time budget.  In \mnist models that are less complex, \tool exhibits comparable performance with the baseline approaches. These results show the efficiency of \tool in handling complex models.


In the scatter plots of Fig.~\ref{fig:speedup}, $x$-axis denotes the time cost for individual problems of \tool, and $y$-axis denotes the speedup (i.e., $\frac{T_{\bab-\text{baseline}}}{T_{\tool}}$) of \tool over \bab-baseline. It reveals significant performance improvement across different models, we can observe many instances for which \tool outperforms \bab-baseline. In \mnist, the speedups are around 1-5 times, and in \cifar, the speedups are around 20-80 times. 
In many cases, while \bab-baseline struggles, \tool manages to solve the problems very efficiently.

\begin{figure*}[!tb]
    \centering
       \begin{subfigure}[b]{0.31\linewidth}
        \centering
        \includegraphics[width=\linewidth]{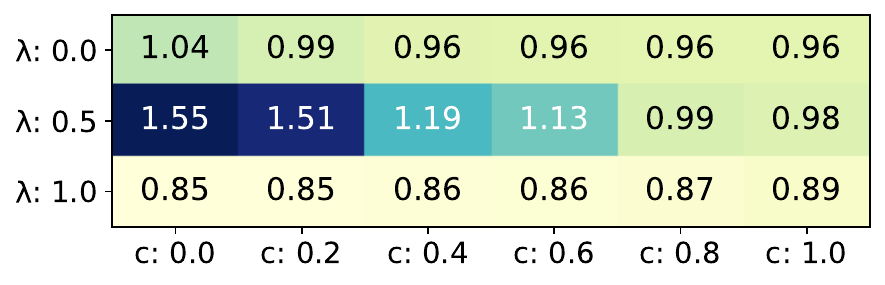}
        \caption{Avg. speedup (w.r.t. \bab-baseline)}
        \label{fig:heatmap1}
    \end{subfigure}
    \hfill
    \begin{subfigure}[b]{0.31\linewidth}
        \centering
        \includegraphics[width=\linewidth]{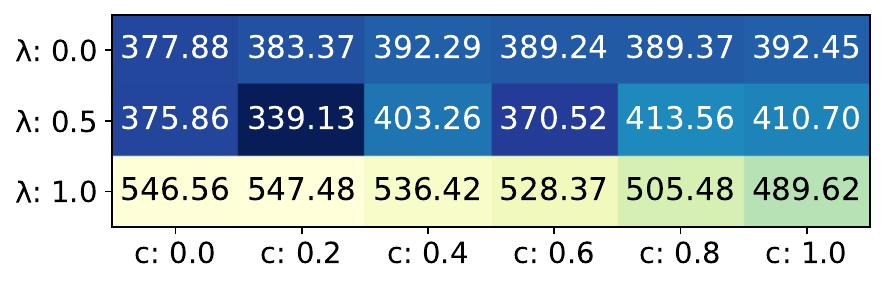}
        \caption{Avg. time (secs)}
        \label{fig:heatmap2}
    \end{subfigure}
    \hfill
    \begin{subfigure}[b]{0.31\linewidth}
        \centering
        \includegraphics[width=\linewidth]{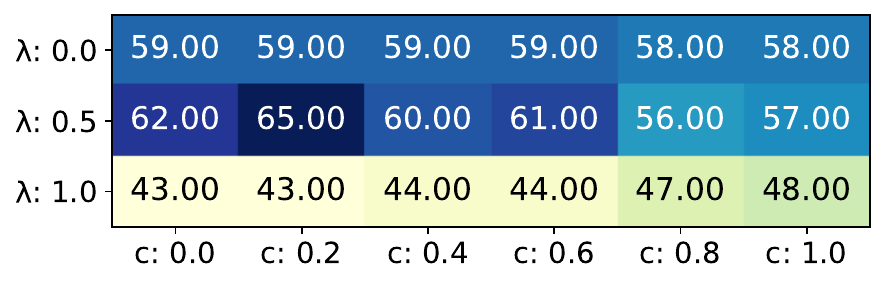}
        \caption{Number of solved problems}
        \label{fig:heatmap3}
    \end{subfigure}
    \caption{RQ2 -- Impact of  hyperparameter selection across different $\lambda$ and $c$. A darker cell implies a better performance.}
    \label{fig:heatmaps}
\end{figure*}

\myparagraph{RQ2: Impacts of  hyperparameter selection on \tool}

Figures \ref{fig:heatmap1}, \ref{fig:heatmap2}, and \ref{fig:heatmap3} illustrate the impact of hyperparameters ($\lambda$ in counterexample potentiality, Def.~\ref{def:cepo} and $c$ in UCB1, Line~\ref{line:UCB1} of Alg.~\ref{alg:mcts_nn_verify}) to the performance of \tool.  

Regarding the selection of $\lambda$, all the plots show that $\lambda = 0.5$ is the best choice. The value $\lambda = 0.5$ aims at a balance between the two node attributes in counterexample potentiality, and the results show that the likelihood of counterexample existence is closely correlated to both of the two attributes.

The hyperparameter $c$ is crucial in MCTS~\cite{zhang2018two}, which decides the extent to which it favors ``exploitation'' or ``exploration''. While in Fig.~\ref{fig:heatmap1} $c=0.2$ (i.e., balance between ``exploitation'' and ``exploration'' to some extent) is a bit weaker than $c = 0$ (i.e., pure exploitation), in Fig.~\ref{fig:heatmap2} and Fig.~\ref{fig:heatmap3}, $c = 0.2$ is the best performer. This indicates that, 1) our counterexample potentiality is effective as a search guidance, so even pure exploitation performs well; 2) pure exploitation may be superior in individual problems, but not as good as the balanced strategy in average performance. So, it is meaningful to strike a balance between ``exploitation'' and ``exploration'' in our MCTS.




\begin{figure}[!tb]
    \centering
    \begin{subfigure}[b]{0.35\linewidth}
        \includegraphics[width=\textwidth]{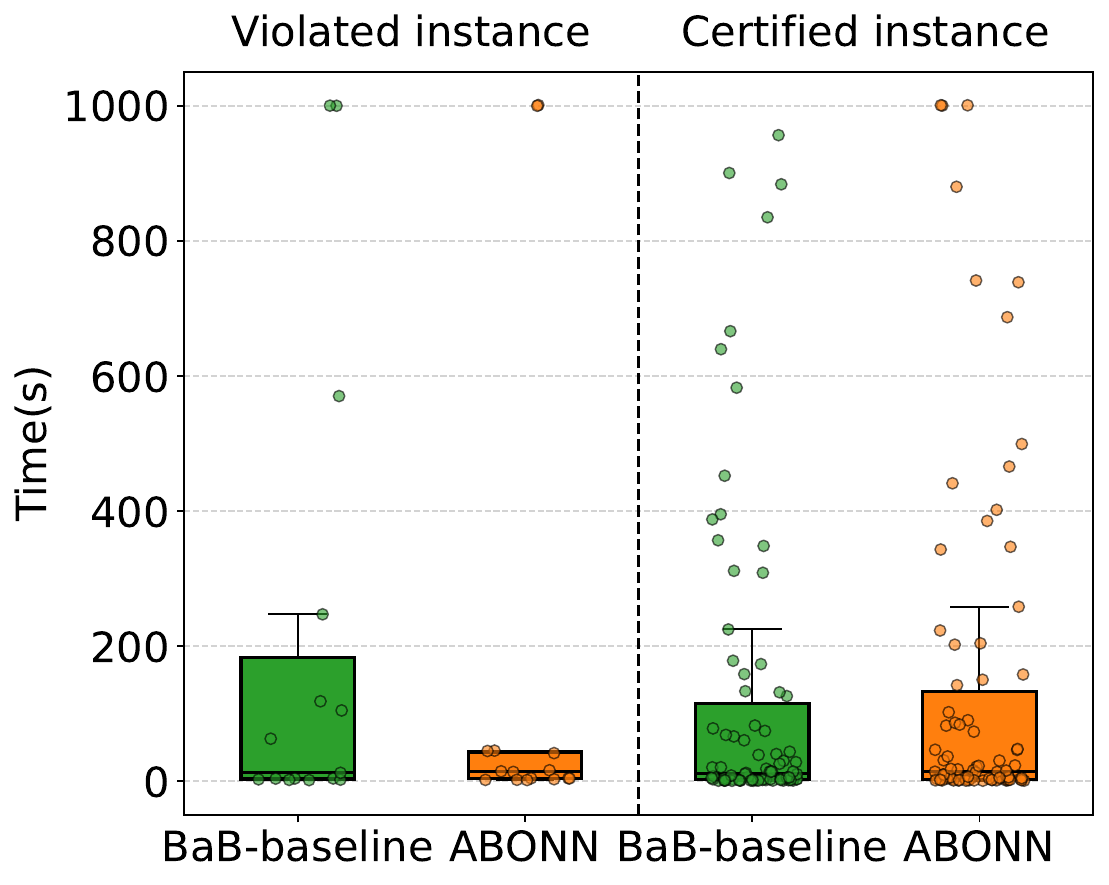}
        \caption{$\mnist_{\ltwo}$}
        \label{fig:boxmnist2}
    \end{subfigure}
    \quad
    \begin{subfigure}[b]{0.35\linewidth}
        \includegraphics[width=\textwidth]{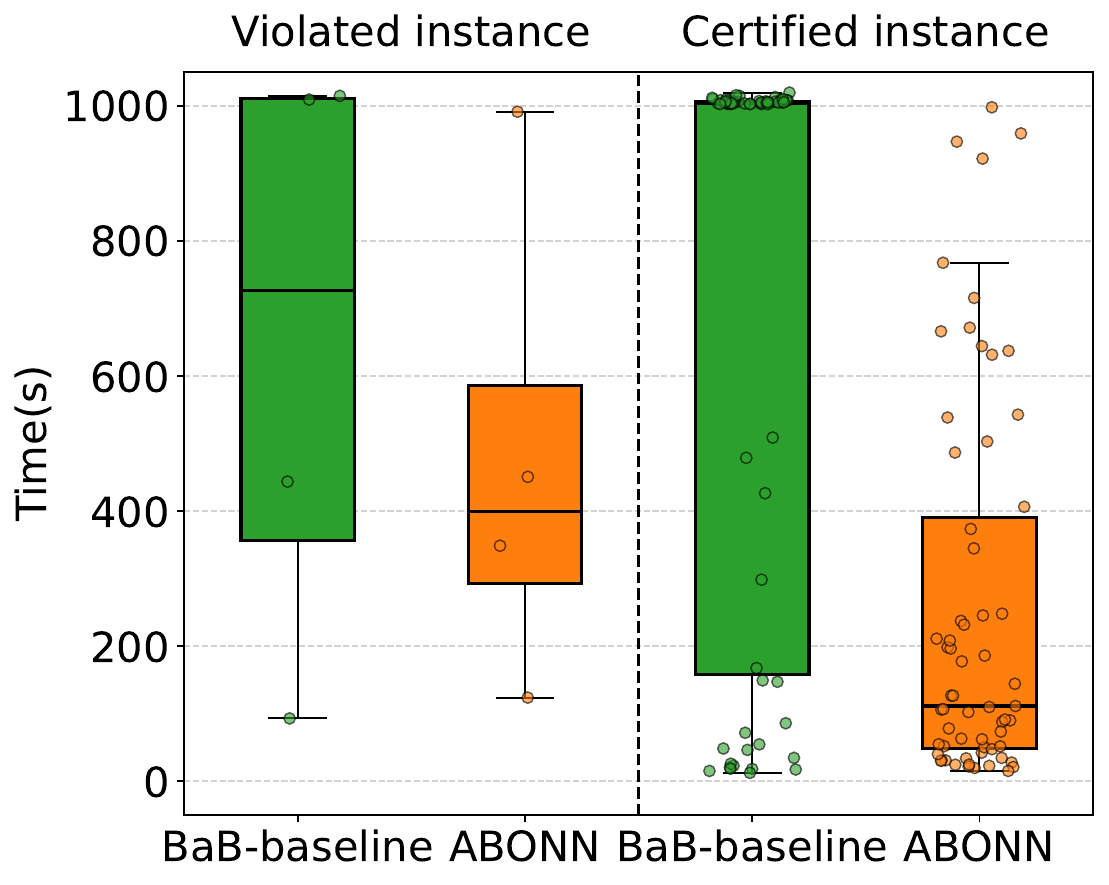}
        \caption{$\cifar_{{\deep}}$}
        \label{fig:sub_deep}
    \end{subfigure}
    \caption{RQ3 -- Comparison between \bab-baseline and \tool for violated and certified verification problems}
    \label{fig:cmp_vio_safe}
\end{figure}

\myparagraph{RQ3: Comparison between \bab-baseline and \tool for violated and certified verification problems}


The box plots in Fig.~\ref{fig:cmp_vio_safe} shows the breakdown of the verification time, respectively for violated problems and for certified problems, of $\mnist_\ltwo$ and $\cifar_\deep$.

In Fig.~\ref{fig:boxmnist2}, \tool 
exhibits a smaller interquartile range than \bab-baseline for violated instances. For certified instances, the two approaches are almost the same. This certifies that \tool is indeed superior in finding counterexamples compared to \bab-baseline, and so it outperforms \bab-baseline when dealing with violated verification problems. For certified instances, it performs similarly to \bab-baseline, which is also expected.



In Fig.~\ref{fig:sub_deep}, similarly, \tool exhibits its superiority in handling violated verification problems, evidently outperforming \bab-baseline. While \bab-baseline used up the time budget ($1000s$) for many problems, \tool solves them much more efficiently. 
Surprisingly, \tool also demonstrates  outperformance in certified instances, which means that it manages to verify the problems by visiting less sub-problems. This should result from a mutual influence between the ReLU selection heuristic~\cite{henriksen2021deepsplit} and our adaptive tree exploration strategy, namely, our strategy manages to lead the heuristic to a better ReLU selection. In future work, we will delve more into the impact of our strategy to ReLU selection.   

\section{Related Work}\label{sec:related}
\myparagraph{Neural Network Verification} Neural network verification has been extensively studied in the past decade~\cite{tjeng2018evaluating,katz2017reluplex,ehlers2017formal,huang2017safety,singh2019abstract,muller2020neural,wang2018efficient, shi2022efficiently} and approximated methods are often preferable thanks to their efficiency~\cite{anderson1811strong,tjandraatmadja2020convex, singh2019beyond,muller2022prima,muller2021precise,raghunathan2018semidefinite}. In particular, there is a line of work~\cite{yang2021improving,ostrovsky2022abstraction, zhao2022cleverest} that exploits information from (spurious) counterexamples to refine the abstraction (i.e., CEGAR~\cite{clarke2000counterexample}). In contrast, we do not use counterexamples to refine the approximation of verifiers, but we explore the sub-problem space of \bab guided by the possibility of finding counterexamples.

\myparagraph{Branching Strategies in \bab} Many studies~\cite{bunel2020branch,henriksen2021deepsplit,de2021improved,shi2024neural,lu2019neural,ferrari2022complete} aim to optimize the branching strategy (i.e., $\reluHeuristic$ in Alg.~\ref{alg:mcts_nn_verify}) to pursue better abstraction refinement. In contrast, our approach is orthogonal to that line of works, and we can adopt any of those strategies in our algorithm. 
Essentially, we change the way of tree growth in \bab by favoring those branches that are more ``promising'', rather than the way of problem splitting. 




\myparagraph{Testing and Attacks} 
Testing and attacks aim to efficiently generate counterexamples to fool the network, and they have been extensively studied~\cite{pei2017deepxplore, odena2019tensorfuzz, goodfellow2015explaining,andriushchenko2020square,xie2019improving,zhang2022branch,carlini2017towards}.
Although these approaches are efficient, they cannot provide rigorous guarantee on the quality of the networks, even if they fail to detect counterexamples. 
In comparison, in line with \bab, our approach is a verification approach that can provide rigorous proofs about specification satisfaction, after verifying all the sub-problems.






\section{Conclusion and Future Work}
We propose a neural network verification approach that features adaptive exploration of the sub-problems produced by \bab. \tool is guided by \emph{counterexample potentiality}, so we can efficiently find the sub-problems that contain counterexamples. Experimental evaluation demonstrates the superiority of our proposed approach in efficiency over existing baselines.


As future work, we aim to investigate how our approach can be used to improve ReLU selection heuristics, such as DeepSplit~\cite{henriksen2021deepsplit}, such that we can further accelerate verification.

\section*{Acknowledgements}
This research is supported by JSPS KAKENHI Grant No. JP23H03372, No. JP23K16865, JST-Mirai Grant No. JPMJMI20B8, and Australian Research Grant FT220100391.



\bibliographystyle{IEEEtran}
\bibliography{date25}

\begin{thebibliography}{10}
\providecommand{\url}[1]{#1}
\csname url@samestyle\endcsname
\providecommand{\newblock}{\relax}
\providecommand{\bibinfo}[2]{#2}
\providecommand{\BIBentrySTDinterwordspacing}{\spaceskip=0pt\relax}
\providecommand{\BIBentryALTinterwordstretchfactor}{4}
\providecommand{\BIBentryALTinterwordspacing}{\spaceskip=\fontdimen2\font plus
\BIBentryALTinterwordstretchfactor\fontdimen3\font minus \fontdimen4\font\relax}
\providecommand{\BIBforeignlanguage}[2]{{%
\expandafter\ifx\csname l@#1\endcsname\relax
\typeout{** WARNING: IEEEtran.bst: No hyphenation pattern has been}%
\typeout{** loaded for the language `#1'. Using the pattern for}%
\typeout{** the default language instead.}%
\else
\language=\csname l@#1\endcsname
\fi
#2}}
\providecommand{\BIBdecl}{\relax}
\BIBdecl

\bibitem{liu2021algorithms}
C.~Liu, T.~Arnon, C.~Lazarus, C.~Strong, C.~Barrett, M.~J. Kochenderfer \emph{et~al.}, ``Algorithms for verifying deep neural networks,'' \emph{Foundations and Trends{\textregistered} in Optimization}, vol.~4, no. 3-4, pp. 244--404, 2021.

\bibitem{goodfellow2015explaining}
I.~J. Goodfellow, J.~Shlens, and C.~Szegedy, ``Explaining and harnessing adversarial examples,'' in \emph{3rd Int. Conf. on Learning Representations (ICLR'15)}.\hskip 1em plus 0.5em minus 0.4em\relax San Diego, CA, United States: Int. Conf. on Learning Representations, ICLR, 2015.

\bibitem{muller2022third}
M.~N. M{\"u}ller, C.~Brix, S.~Bak, C.~Liu, and T.~T. Johnson, ``3rd international verification of neural networks competition ({VNN-COMP} 2022): Summary and results,'' \emph{arXiv preprint arXiv:2212.10376}, 2022.

\bibitem{cheng2017maximum}
C.-H. Cheng, G.~N{\"u}hrenberg, and H.~Ruess, ``Maximum resilience of artificial neural networks,'' in \emph{Automated Technology for Verification and Analysis}, D.~D'Souza and K.~Narayan~Kumar, Eds.\hskip 1em plus 0.5em minus 0.4em\relax Springer Int. Publishing, 2017, pp. 251--268.

\bibitem{tjeng2018evaluating}
V.~Tjeng, K.~Y. Xiao, and R.~Tedrake, ``Evaluating robustness of neural networks with mixed integer programming,'' in \emph{Int. Conf. on Learning Representations}, 2018.

\bibitem{singh2018fast}
G.~Singh, T.~Gehr, M.~Mirman, M.~P{\"u}schel, and M.~Vechev, ``Fast and effective robustness certification,'' \emph{Advances in neural information processing systems}, vol.~31, 2018.

\bibitem{singh2019abstract}
G.~Singh, T.~Gehr, M.~P{\"u}schel, and M.~Vechev, ``An abstract domain for certifying neural networks,'' \emph{ACM on Programming Languages}, vol.~3, no. POPL, pp. 1--30, 2019.

\bibitem{zhang2018efficient}
H.~Zhang, T.-W. Weng, P.-Y. Chen, C.-J. Hsieh, and L.~Daniel, ``Efficient neural network robustness certification with general activation functions,'' \emph{Advances in Neural Information Processing Systems}, vol.~31, 2018.

\bibitem{wong2018provable}
E.~Wong and Z.~Kolter, ``Provable defenses against adversarial examples via the convex outer adversarial polytope,'' in \emph{Int. Conf. on Machine Learning}.\hskip 1em plus 0.5em minus 0.4em\relax PMLR, 2018, pp. 5286--5295.

\bibitem{bunel2020branch}
R.~Bunel, P.~Mudigonda, I.~Turkaslan, P.~Torr, J.~Lu, and P.~Kohli, ``Branch and bound for piecewise linear neural network verification,'' \emph{Journal of Machine Learning Research}, vol.~21, no. 2020, 2020.

\bibitem{browne2012survey}
C.~B. Browne, E.~Powley, D.~Whitehouse, S.~M. Lucas, P.~I. Cowling, P.~Rohlfshagen, S.~Tavener, D.~Perez, S.~Samothrakis, and S.~Colton, ``A survey of {Monte Carlo} tree search methods,'' \emph{IEEE Transactions on Computational Intelligence and AI in games}, vol.~4, no.~1, pp. 1--43, 2012.

\bibitem{ugare2023incremental}
S.~Ugare, D.~Banerjee, S.~Misailovic, and G.~Singh, ``Incremental verification of neural networks,'' \emph{ACM on Programming Languages}, vol.~7, no. PLDI, pp. 1920--1945, 2023.

\bibitem{wang2021beta}
S.~Wang, H.~Zhang, K.~Xu, X.~Lin, S.~Jana, C.-J. Hsieh, and J.~Z. Kolter, ``Beta-crown: Efficient bound propagation with per-neuron split constraints for neural network robustness verification,'' \emph{Advances in Neural Information Processing Systems}, vol.~34, pp. 29\,909--29\,921, 2021.

\bibitem{henriksen2021deepsplit}
P.~Henriksen and A.~Lomuscio, ``Deepsplit: An efficient splitting method for neural network verification via indirect effect analysis.'' in \emph{IJCAI}, 2021, pp. 2549--2555.

\bibitem{de2021improved}
A.~De~Palma, R.~Bunel, A.~Desmaison, K.~Dvijotham, P.~Kohli, P.~H. Torr, and M.~P. Kumar, ``Improved branch and bound for neural network verification via lagrangian decomposition,'' \emph{arXiv preprint arXiv:2104.06718}, 2021.

\bibitem{singh2018deepz}
G.~Singh, T.~Gehr, M.~Mirman, M.~P\"{u}schel, and M.~Vechev, ``Fast and effective robustness certification,'' in \emph{Advances in Neural Information Processing Systems}, vol.~31, 2018.

\bibitem{zhang2018two}
Z.~Zhang, G.~Ernst, S.~Sedwards, P.~Arcaini, and I.~Hasuo, ``Two-layered falsification of hybrid systems guided by monte carlo tree search,'' \emph{IEEE Transactions on Computer-Aided Design of Integrated Circuits and Systems}, vol.~37, no.~11, pp. 2894--2905, 2018.

\bibitem{katz2017reluplex}
G.~Katz, C.~Barrett, D.~L. Dill, K.~Julian, and M.~J. Kochenderfer, ``Reluplex: An efficient {SMT} solver for verifying deep neural networks,'' in \emph{Computer Aided Verification}, R.~Majumdar and V.~Kun{\v{c}}ak, Eds.\hskip 1em plus 0.5em minus 0.4em\relax Springer Int. Publishing, 2017, pp. 97--117.

\bibitem{ehlers2017formal}
R.~Ehlers, ``Formal verification of piece-wise linear feed-forward neural networks,'' in \emph{Automated Technology for Verification and Analysis: 15th Int. Symp., ATVA 2017, Proceedings 15}.\hskip 1em plus 0.5em minus 0.4em\relax Springer, Oct. 2017, pp. 269--286.

\bibitem{huang2017safety}
X.~Huang, M.~Kwiatkowska, S.~Wang, and M.~Wu, ``Safety verification of deep neural networks,'' in \emph{Computer Aided Verification: 29th Int. Conf., CAV 2017, Part I 30}.\hskip 1em plus 0.5em minus 0.4em\relax Springer, July 2017, pp. 3--29.

\bibitem{muller2020neural}
C.~M{\"u}ller, G.~Singh, M.~P{\"u}schel, and M.~T. Vechev, ``Neural network robustness verification on gpus,'' \emph{CoRR, abs/2007.10868}, 2020.

\bibitem{wang2018efficient}
S.~Wang, K.~Pei, J.~Whitehouse, J.~Yang, and S.~Jana, ``Efficient formal safety analysis of neural networks,'' \emph{Advances in neural information processing systems}, vol.~31, 2018.

\bibitem{shi2022efficiently}
Z.~Shi, Y.~Wang, H.~Zhang, J.~Z. Kolter, and C.-J. Hsieh, ``Efficiently computing local lipschitz constants of neural networks via bound propagation,'' \emph{Advances in Neural Information Processing Systems}, vol.~35, pp. 2350--2364, 2022.

\bibitem{anderson1811strong}
R.~Anderson, J.~Huchette, C.~Tjandraatmadja, and J.~Vielma, ``Strong convex relaxations and mixed-integer programming formulations for trained neural networks (2018),'' 1811.

\bibitem{tjandraatmadja2020convex}
C.~Tjandraatmadja, R.~Anderson, J.~Huchette, W.~Ma, K.~K. Patel, and J.~P. Vielma, ``The convex relaxation barrier, revisited: Tightened single-neuron relaxations for neural network verification,'' \emph{Advances in Neural Information Processing Systems}, vol.~33, pp. 21\,675--21\,686, 2020.

\bibitem{singh2019beyond}
G.~Singh, R.~Ganvir, M.~P{\"u}schel, and M.~Vechev, ``Beyond the single neuron convex barrier for neural network certification,'' \emph{Advances in Neural Information Processing Systems}, vol.~32, 2019.

\bibitem{muller2022prima}
M.~N. M{\"u}ller, G.~Makarchuk, G.~Singh, M.~P{\"u}schel, and M.~Vechev, ``Prima: general and precise neural network certification via scalable convex hull approximations,'' \emph{ACM on Programming Languages}, vol.~6, no. POPL, pp. 1--33, 2022.

\bibitem{muller2021precise}
------, ``Precise multi-neuron abstractions for neural network certification,'' \emph{arXiv preprint arXiv:2103.03638}, 2021.

\bibitem{raghunathan2018semidefinite}
A.~Raghunathan, J.~Steinhardt, and P.~S. Liang, ``Semidefinite relaxations for certifying robustness to adversarial examples,'' \emph{Advances in neural information processing systems}, vol.~31, 2018.

\bibitem{yang2021improving}
P.~Yang, R.~Li, J.~Li, C.-C. Huang, J.~Wang, J.~Sun, B.~Xue, and L.~Zhang, ``Improving neural network verification through spurious region guided refinement,'' in \emph{Int. Conf. on Tools and Algorithms for the Construction and Analysis of Systems}.\hskip 1em plus 0.5em minus 0.4em\relax Springer, 2021, pp. 389--408.

\bibitem{ostrovsky2022abstraction}
M.~Ostrovsky, C.~Barrett, and G.~Katz, ``An abstraction-refinement approach to verifying convolutional neural networks,'' in \emph{International Symposium on Automated Technology for Verification and Analysis}.\hskip 1em plus 0.5em minus 0.4em\relax Springer, 2022, pp. 391--396.

\bibitem{zhao2022cleverest}
Z.~Zhao, Y.~Zhang, G.~Chen, F.~Song, T.~Chen, and J.~Liu, ``Cleverest: accelerating cegar-based neural network verification via adversarial attacks,'' in \emph{International Static Analysis Symposium}.\hskip 1em plus 0.5em minus 0.4em\relax Springer, 2022, pp. 449--473.

\bibitem{clarke2000counterexample}
E.~Clarke, O.~Grumberg, S.~Jha, Y.~Lu, and H.~Veith, ``Counterexample-guided abstraction refinement,'' in \emph{Computer Aided Verification: 12th International Conference, CAV 2000, Chicago, IL, USA, July 15-19, 2000. Proceedings 12}.\hskip 1em plus 0.5em minus 0.4em\relax Springer, 2000, pp. 154--169.

\bibitem{shi2024neural}
Z.~Shi, Q.~Jin, Z.~Kolter, S.~Jana, C.-J. Hsieh, and H.~Zhang, ``Neural network verification with branch-and-bound for general nonlinearities,'' \emph{arXiv preprint arXiv:2405.21063}, 2024.

\bibitem{lu2019neural}
J.~Lu and M.~P. Kumar, ``Neural network branching for neural network verification,'' \emph{arXiv preprint arXiv:1912.01329}, 2019.

\bibitem{ferrari2022complete}
C.~Ferrari, M.~N. Muller, N.~Jovanovic, and M.~Vechev, ``Complete verification via multi-neuron relaxation guided branch-and-bound,'' \emph{arXiv preprint arXiv:2205.00263}, 2022.

\bibitem{pei2017deepxplore}
K.~Pei, Y.~Cao, J.~Yang, and S.~Jana, ``Deepxplore: Automated whitebox testing of deep learning systems,'' in \emph{26th Symp. on Operating Systems Principles}, 2017, pp. 1--18.

\bibitem{odena2019tensorfuzz}
A.~Odena, C.~Olsson, D.~Andersen, and I.~Goodfellow, ``Tensorfuzz: Debugging neural networks with coverage-guided fuzzing,'' in \emph{International Conference on Machine Learning}.\hskip 1em plus 0.5em minus 0.4em\relax PMLR, 2019, pp. 4901--4911.

\bibitem{andriushchenko2020square}
M.~Andriushchenko, F.~Croce, N.~Flammarion, and M.~Hein, ``Square attack: a query-efficient black-box adversarial attack via random search,'' in \emph{European conference on computer vision}.\hskip 1em plus 0.5em minus 0.4em\relax Springer, 2020, pp. 484--501.

\bibitem{xie2019improving}
C.~Xie, Z.~Zhang, Y.~Zhou, S.~Bai, J.~Wang, Z.~Ren, and A.~L. Yuille, ``Improving transferability of adversarial examples with input diversity,'' in \emph{Proceedings of the IEEE/CVF conference on computer vision and pattern recognition}, 2019, pp. 2730--2739.

\bibitem{zhang2022branch}
H.~Zhang, S.~Wang, K.~Xu, Y.~Wang, S.~Jana, C.-J. Hsieh, and Z.~Kolter, ``A branch and bound framework for stronger adversarial attacks of relu networks,'' in \emph{International Conference on Machine Learning}.\hskip 1em plus 0.5em minus 0.4em\relax PMLR, 2022, pp. 26\,591--26\,604.

\bibitem{carlini2017towards}
N.~Carlini and D.~Wagner, ``Towards evaluating the robustness of neural networks,'' in \emph{2017 ieee symposium on security and privacy (sp)}.\hskip 1em plus 0.5em minus 0.4em\relax Ieee, 2017, pp. 39--57.

\end{thebibliography}
\end{document}